\documentclass[hf]{ceurart}

\usepackage{amssymb}
\usepackage{amsmath}
\usepackage{comment}
\usepackage{array, makecell} 
\usepackage{tikz}
\usepackage{pgfplots}
\usepackage{pgfplotstable}

\pgfplotsset{compat=1.8}
\usepackage{booktabs}
\usepackage{csquotes}

\begin{document}

\copyrightyear{2021}
\copyrightclause{Copyright for this paper by its authors.
  Use permitted under Creative Commons License Attribution 4.0
  International (CC BY 4.0).}

\conference{ISWC 2021 Workshop DL4KG}

\title{Generating Table Vector Representations}

\author[1,2]{Aneta Koleva}
\address[1]{Siemens, Otto-Hahn-Ring 6, 81739 Munich, Germany}
\address[2]{Ludwig Maximilian University of Munich, Geschwister-Scholl-Platz 1,
80539 Munich, Germany}

\author[1]{Martin Ringsquandl}
\author[1]{Mitchell Joblin}
\author[1,2]{Volker Tresp}

\ead{firstname.lastname@siemens.com}

\begin{abstract}
  High-quality Web tables are rich sources of information that can be used to populate Knowledge Graphs~(KG). The focus of this paper is an evaluation of methods for table-to-class annotation, which is a sub-task of Table Interpretation (TI). We provide a formal definition for table classification as a machine learning task. We propose an experimental setup and we evaluate $5$ fundamentally different approaches to find the best method for generating vector table representations. Our findings indicate that although transfer learning methods achieve high F$1$ score on the table classification task, dedicated table encoding models are a promising direction as they appear to capture richer semantics. 
\end{abstract}

\begin{keywords}
  table interpretation, table classification, representation learning.
\end{keywords}

\maketitle

\section{Introduction}
Tabular data is one of the most prevalent data representations. The effort by Cafarella \cite{webTables1}, known as WebTables, identified and extracted more than 200 million high-quality tables from HTML pages. The availability of such large corpus of structured data initiated several directions of research related to the different applications of tabular data such as: table search \cite{recoveringSem}, table improvement \cite{tableImprove}, question answering \cite{tableQuery}, and semantic annotation of columns \cite{semAnnotation}. 
As a result of the increasing adoption of KGs, which are often populated from tabular data, the task of aligning tables with KGs, also referred to as table interpretation (TI), has become a highly relevant task. In contrast to information extraction from unstructured documents, TI should leverage the explicit relational structure. The unique table structure with rows and columns of cells and other \textit{metadata} can be exploited for discovery and disambiguation of the meaning captured in the table. 
The task of TI entails three different sub-tasks.
The first sub-task, which is the focus in this paper, is the classification of tables according to classes in a given KG schema. The second sub-task is related to linking rows from tables to existing entities in the KG. The annotation of columns as entity attributes and the discovery of binary relations between columns is the third sub-task of TI. While there have been several works focusing on the row-to-entity \cite{instanceMatch1, instanceMatch2, instanceMatch3}, and column-to-attribute sub-tasks \cite{semAnnotation,Limaye},
the task of linking a table to a class has been neglected. However, in the case of \textit{entity tables}, where one column (the \textit{core column}) is associated to the name of the entity and the remaining columns are attributes of this entity, discovering the class of the table as a first step can greatly improve the solving of the other two sub-tasks. It is often the case that the column names are missing or incorrect, therefore finding the name of the core column does not imply finding the class of the table.
Moreover, when two tables have the same column names and similar content (e.g., one table of class \textit{Country} and one of class \textit{City}), it is not trivial to disambiguate the entities and column types based only on the table content.
Once a table has been \textit{interpreted}, its content can be used for extracting new triples for enriching the KG, a task known as \textit{KG completion}, or for extracting missing facts for the KG, which is the task of \textit{slot-filling}. 

Due to the inherent scarcity of labelled data for the first sub-task (class-annotated tables), a table classification model must either be of low complexity (few parameters) or 
leverage pre-trained models. Using pre-trained models in TI has been studied only to a very limited extend. Hence, we explore two promising directions for making learning-based approaches more efficient: (a) by using transfer learning, (b) by considering additional inductive biases that are unique to tabular data representations.

We propose an experimental setup with the intention of
finding the best method for generating a representation which captures the information from the table but also the row and column structure, so that it can be later used towards solving the remaining sub-tasks of TI: row-to-entity linking, column type annotation and relation extraction. 
We are interested in understanding how pre-trained language models, such as BERT \cite{bert}, and their dedicated table-based counterparts, for instance TaBERT \cite{tabert}, can be utilized for generating vector representation for table.  
Surprisingly, our experiments show that a transfer learning method with a rich vocabulary of pre-trained word embeddings achieves similar F$1$ score compared to more sophisticated  pre-trained language models (LM). Another interesting finding is that the inductive bias for tabular structure in the LM pre-trained on tabular data does not bring beneficial impact to a text pre-trained LM. However, the classification confusion matrix for this method, gives an insight to the miss-classifications being justifiable and reasonable.
Our main contributions~are:
\begin{itemize}
    \item A formal definition of table classification as a machine learning task and a protocol for evaluating performance on this task.
    \item A setup for table encoding using 5 fundamentally different approaches covering a spectrum of paradigms from general purpose document encoders to specialized pre-trained models designed for tabular data.
    \item An extensive empirical evaluation of the different approaches. 

\end{itemize}
\section{Background} \label{RW}
In this section, we review  prior work related to solving the different sub-tasks of TI. We also give a short overview of methods for generating vector representations of tables.
\paragraph{\textbf{Table Interpretation}} 
The three sub-tasks of TI were first introduced in the paper by Ritze et al. \cite{t2d}. That paper also introduced the T2K Matcher, a method for iterative value-based matching, which solves the TI tasks by matching values from the tables to values of retrieved candidates from the KG. More recent work by Limaye et al. \cite{Limaye} proposed a probabilistic graphical method which attempts to jointly solve the two sub-tasks of finding entity-to-row and column-to-attribute alignments. Deng et al. \cite{tab2vec} exploited word embeddings for representing the contents of tables and utilized them for the discovery of new entities. The SemTab challenge \cite{semtab} has also motivated new approaches \cite{semtab2,semtab1}. However, the task of table-to-class annotation is not part of this challenge.
\paragraph{\textbf{Table classification}} 
To  the  best  of  our  knowledge, the T2K Matcher is the only existing method for solving the table-to-class task. Namely, the class of the table is chosen by ranking the sum of the similarity scores of the column-to-property correspondences aggregated per class. Since this method requires querying of the KG for candidate retrieval and first solving the column-to-property alignment in  order  to  find  the  correct  class  of  a  table,  we  do  not  consider  it  during  our experiments. In contrast to the T2K Matcher, we consider a \textit{closed book} scenario, where the instances of the KG are not available, only the classes in~the~KG~schema.
\paragraph{\textbf{Representation Learning on Tables}} 
Based on powerful LM, dedicated \emph{deep learning} models have recently been proposed  to exploit tabular data structures, e.g., in table-based question answering \cite{tableQuery, turl} and KG completion from tables \cite{novelFacts}. 
One benefit from using pre-trained LM is that they can handle synonyms well, e.g., the  abbreviation of New York as NY, which are frequently occurring in tables because of the innate limitation of the cells. The other benefit is that, due to the exposure to large textual corpora during the pre-training phase, the LM can \textit{store} implicit information learned from the data whilst pre-training, in the form of model parameters \cite{nlp}. 
TaBERT \cite{tabert} by Yin et al. is a novel model which was pre-trained to jointly learn representation of a natural language question, called \textit{utterance}, and tables. An example of utterance for the entity table shown in Figure \ref{fig:tab_encoder} is the question: \textit{How much is the population of New York?}. During encoding, instead of using the full table, TaBERT  samples  $1$ or $3$ rows, referred to as \textit{content snapshot}. First, each row from the snapshot, concatenated with the utterance, is encoded by BERT \cite{bert}. Second, the encoding of the rows are stacked and in order to generate vector representations for each of the columns, a vertical self-attention mechanism is used.  Finally, representation for the table is generated by pooling the column representations. Similar work is the method TAPAS by Herzig et al. \cite{tapas}, which is also pre-trained on tables and text segments.
Ding et al. proposed TURL \cite{turl} as a framework for pre-training, also on tabular data, which uses the same objectives as TaBERT for learning representations of the content of the tables. Additionally, they proposed task-specific fine-tuning on the framework for solving the row-to-entity and column-to-attribute annotation. Wang et al. \cite{TCN} presented a novel method which exploits information within one table but also aggregates the contextual information shared across similar tables in order to generate a vector representation that can be used for column-to-class annotation and relation prediction tasks.

\section{Problem Description}
We focus on the task of table-to-class annotation. The task has been introduced together with the two other TI sub-tasks in \cite{t2d}, however without a formal definition. The goal of the table-to-class annotation is to label a table with its corresponding class according to the given KG schema. We now provide a definition of this task as a machine learning task.

An entity table $\boldsymbol{T}_i$ is a $N_i \times M_i$ matrix
where $N_i$ and $M_i$ are the number of rows and columns of the table $\boldsymbol{T}_i$. 
Each element of the matrix $\boldsymbol{T}_i$, $r_{n,m}^i$, contains one or more tokens, where each token is a sequence of characters. 
We denote with $r_{n,*}^i$ and $r_{*,m}^i$ the $n$-th row and the $m$-th column of the matrix $\boldsymbol{T}_i$ respectively. The header of the table is the first row $H_i=r_{0,*}^i$. The content of the table are the rows $r_{1,*}^i, r_{2,*}^i, \dots, r_{N,*}^i$.

Let $\mathcal{D}= \{(\boldsymbol{T}_1, c_i), \dots, (\boldsymbol{T}_l, c_i)\}$ be the set of labeled tables with $l$ number of tables,  
and each label $c_i \in C$ is in the set of classes defined in the KG schema $\mathcal{C}= \{c_1, \dots, c_k \}$. A table encoder $E_{\boldsymbol{\omega}}$ is a model, with a parameter vector $\boldsymbol{\omega}$, which encodes each table $E_{\boldsymbol{\omega}} : \{\boldsymbol{T}_i\} \rightarrow \mathbb{R}^d $ to a vector $E_{\boldsymbol{\omega}}(\boldsymbol{T}_i) = \boldsymbol{x}_i$ 
and $\mathcal{X} = \{\boldsymbol{x}_0,\boldsymbol{x}_1, \dots, \boldsymbol{x}_l\} $ is the set of feature vectors for every $\boldsymbol{T}_i \in \mathcal{D}$. 
The final task is to train a classification model $f_{\boldsymbol{\theta}} :\mathbb{R}^d \rightarrow \mathcal{C} $ so that each table vector is assigned to one of the class labels. The problem is defined in the multi-class setting. Formally our setting is $ f_{\boldsymbol{\theta}}  \circ E_{\boldsymbol{\omega}} : \left\{\boldsymbol{T}_i\right\} \rightarrow \mathcal{C}$, where only the parameters $\boldsymbol{\theta}$ are trained on the table classification task, i.e., no gradient updates are performed~on~$\boldsymbol{\omega}$. 


\section{Experiments}
 
\begin{figure}[h!]
    \centering
    \includegraphics[width=0.93\textwidth]{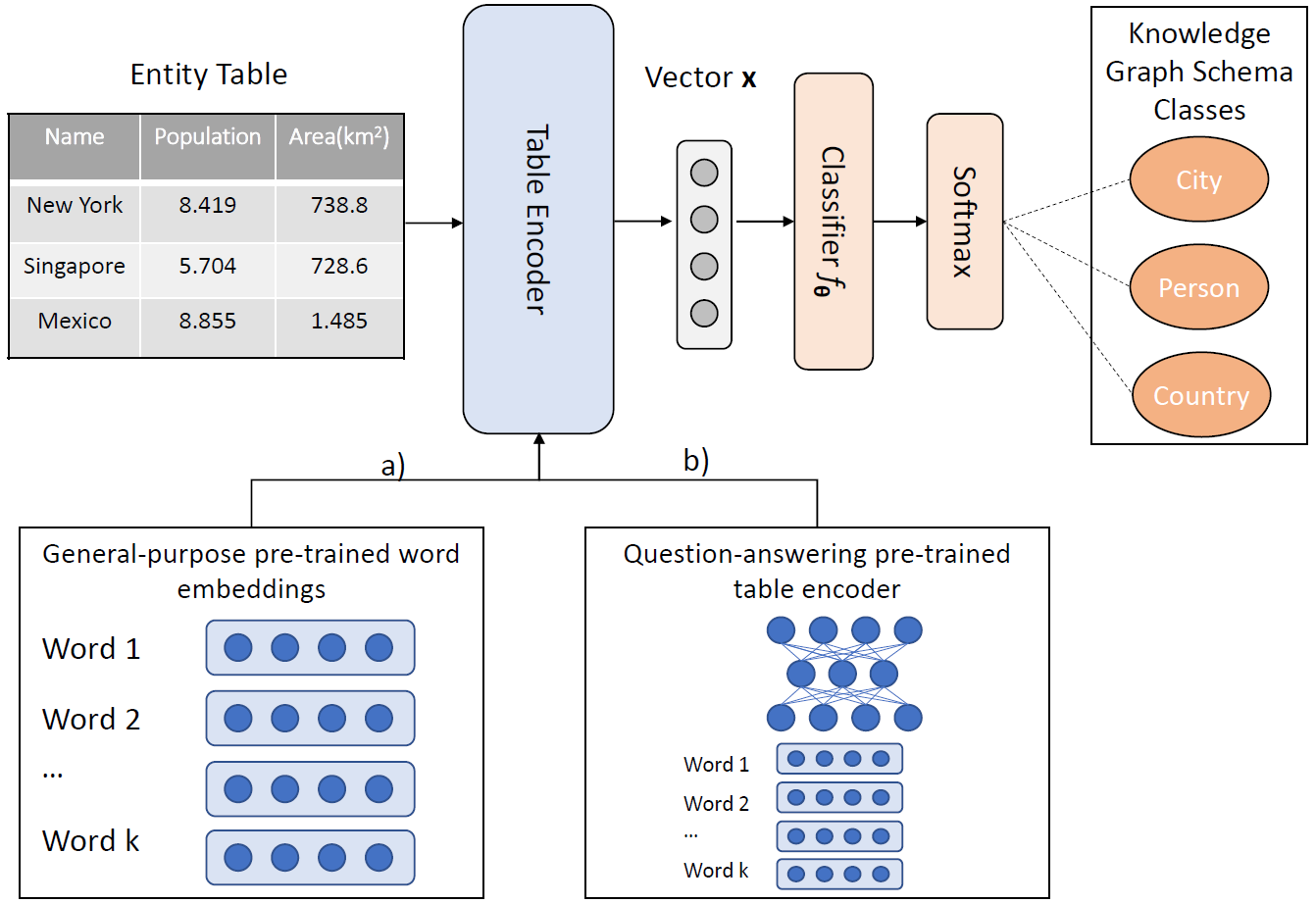}
    \caption{Experimental setup for evaluation of table encoders.}
    \label{fig:tab_encoder}
\end{figure}

Figure \ref{fig:tab_encoder} shows the experimental setup for evaluating different table encoders. Given an entity table, a table encoder generates a high-dimensional vector representation of the table. We then train a classifier on the table-to-class task and evaluate the performance achieved by each of the table encoders. 
We experiment with different types of table encoders, a simple method such as document encoder, transfer learning methods with general-purpose pre-trained word embeddings (Figure \ref{fig:tab_encoder} (a)) and more complex methods which include a LM pre-trained on large textual corpora and an approach for question-answering which has been pre-trained on tabular data (Figure \ref{fig:tab_encoder} (b)). The code for the experiments~is~accessible~online \footnote{https://github.com/anetakoleva/tableClassification}.

\subsection{Dataset}
For evaluation we used the second version of the T2D gold standard dataset~\cite{t2d}, T2Dv2. To the best of our knowledge, the T2D sets are the only publicly available datasets which have been annotated with table-to-class correspondence. The second version of the dataset\footnote{http://webdatacommons.org/webtables/goldstandardV2.html} contains $237$ such annotations. In our experiments, we consider those classes which have at least two tables as representatives.
The resulting dataset contains $223$ tables, each labeled with one of the $27$~unique~classes. The mean of the number of rows in the dataset is $119.2$ and the mean of the number of columns is $7.7$.

\subsection{Models compared}
In the evaluation we used $5$ different models as table encoders, varying from general purpose document encoders to more sophisticated LM, pre-trained on tabular data. 
\paragraph{TF-IDF} or term frequency-inverse document frequency, is a term weighting scheme which generates vector representation for a document based on the frequency of the words in the document. It is the simplest method which we used as a table~encoder.
\paragraph{Spacy} pre-trained word vectors on a text extracted from blogs, news and comments. We used the vectorizer from english-medium sized pipeline\footnote{https://spacy.io/models/en\#en\_core\_web\_md}
which contains vocabulary of size $684 830$.  
\paragraph{Word2Vec} pre-trained word vectors trained with FastText \footnote{https://fasttext.cc/docs/en/pretrained-vectors.html} on a Wikipedia text corpus. The model used for the learning the vectors \cite{w2vec} is an extension of the original word2vec model. It is skip-gram based and trained to learn representations for character n-grams. This model consists of vocabulary of size~$2.5$~million.
\paragraph{BERT} is a widely used, Transformer-based LM \cite{bert}. During the pre-training phase, the model has been exposed to a large corpus of unstructured text with the objective of predicting missing words and prediction of next sentence. This enables the model to learn the correlation of the words and to generate different vector representation for words depending on the context.
\paragraph{TaBERT} is a table encoding method \cite{tabert}, pre-trained on Web tables with the objective to be used in question-answering tasks on  tables. Since the model expects an \textit{utterance}, i.e., a natural language question, as input together with a table, in our experiments we provided an empty space \textquote{ }. We conducted more experiments to evaluate the influence of the utterance on the generated table representation and we discuss these results in Section \ref{results}.

\subsection{Setup}
To systematically evaluate the quality of the representations generated with the different table encoders, we compare their performance on the classification task under different scenarios. It is important to note that we did not train or fine-tune any of the methods for table encoding, i.e., we used them \textit{off-the-shelf}. 
Since the tables can be large, in order to avoid scalability issues, we resort to sampling of rows. Namely, we first shuffle the rows in the tables and then we sample the first $q$ rows. The shuffling of the rows is done only once. For the experiments, we sampled $q \in \{1,3,5,7\}$ rows from each of the tables and used these sampled tables as input to the table encoders. 

When using TF-IDF as table encoder, the input is a set of sequences, where each sequence corresponds to a table from the set of tables $\mathcal{D}$. More formally, a table sequence for table $\boldsymbol{T}_i$ is a sequence of rows $S_{T_i} = (r_{0,*}^i,r_{1,*}^i,\dots, r_{q,*}^i)$, such that $q \in \{1,3,5,7\}$, and the set of sequences is the set $I = \{S_{T_0}, \dots, S_{T_l}\}$. The table encoder TF-IDF transforms the set of table sequences to the set of feature vectors ~$E_{\omega}^{\text{tf-idf}}:~I~\rightarrow~\mathcal{X}$.

Word2Vec and Spacy generate the vector representation for table $\boldsymbol{T}_i$ in $3$ steps. First, the sequence $S_{H_i}$, representing the header of the table $T_{i}$, is encoded as the mean over the word vectors in the sequence $S_{H_i}$, represented as $\boldsymbol{x}_H^i$. Second, the content of the table, is transformed into a table sequence $S_{T_i} = (r_{1,*}^i \dots r_{q,*}^i )$ and encoded as the vector $\boldsymbol{x}_B^i$, which represents the mean over all the word vectors in $S_{T_i}$. Finally, the vector representations for the header and for the table content are concatenated into one vector~$\boldsymbol{x}_{T_i}= \boldsymbol{x}^i_H \| \boldsymbol{x}^i_B$. 

Considering that there is a limit on the length of the sequence that BERT can encode in one step, we used different transformation for the last two methods. BERT encodes each table row by row, i.e, a sequence $S^i_{r_{z,*}}$ is generated for each of the rows $r_{z,*}^i$ of table $\boldsymbol{T}_i$, where $0 \leq z \leq q$. BERT generates row-wise vectors, so for each sequence $S^i_{r_{z,*}}$ the output is a vector $\boldsymbol{x}_{r_{z,*}}$. The vector representation for table $\boldsymbol{T}_i$ is the vector $\boldsymbol{x}_{T_i}$ which is the result of the mean-pooling over the set of the BERT's output vectors $\{\boldsymbol{x}_{r_{0,*}}, \dots, \boldsymbol{x}_{r_{q,*}}\}$ that correspond to the table rows.
In the same manner, the TaBERT model also first generates an encoding for each of the rows of table $\boldsymbol{T}_i$ resulting in a set of vectors. This model uses vertical self-attention focused on the vertically stacked vectors, $\{\boldsymbol{x}_{r_{0,*}}, \dots, \boldsymbol{x}_{r_{q,*}}\}$. Because of the vertically aligned vectors, the output of the model is a column vector representation $\{\boldsymbol{x}_{r_{*,0}}, \dots ,\boldsymbol{x}_{r_{*,M_i}}\}$ for each of the $M_i$ columns in table $\boldsymbol{T}_i$. Finally, we do mean-pooling over the column representations to generate the table encoding $\boldsymbol{x}_{T_i}$. 

We then use the Multi-layer Perceptron (MLP) with one hidden layer of size $500$, the \textit{tanh} activation function and \textit{adam} optimizer as the classifier $f_{\boldsymbol{\theta}}$ from Figure \ref{fig:tab_encoder}. The hyper parameters are chosen after an extensive search and they are fixed for all of the experiments. Since the available dataset is small, instead of splitting it once into a training set and a test set, we use stratified K-fold validation with $K=20$ splits. Considering that the dataset is imbalanced, we report the macro averaged F1 score. The reported scores are the average of the results on the test set after the cross validation. 
To explore the effect of the column names, we also encoded the tables with their column names masked. Specifically, for all of the tables, we substitute their column names with the token [UNK].  


\section{Results} \label{results}
Table \ref{eval_res} shows the macro averaged F$1$ score for the $5$ table encoders on the table classification task under two different settings: (1) given the input tables with the column names and (2) given the input tables with their column names masked ([UNK] token). We report the achieved F$1$ score for the different sizes of the input tables with the number of sampled rows $q$ varying from $1$ row to $7$ rows. The simplest table encoder, TF-IDF achieves the lowest F$1$ score and the score only got lower when the column names of the tables were masked. For the two models with pre-trained word vectors, we observe that the model with the richer vocabulary has higher score. Indeed, the F$1$ score of Word2Vec is comparable with the scores achieved by BERT and TaBERT. In the first setting, when the column names of the tables are visible, there is no significant difference between the scores achieved by BERT and the scores of TaBERT. However, in the setting when the column names are masked, BERT consistently outperforms TaBERT. Interestingly, Word2Vec is the only table encoder that was not affected by the masking of the column names, on the contrary, it achieved better score in the case when $q=3$ and $q=7$ under the second setting compared to the setting when the column names are visible.

\begin{table} 
\caption{Macro-averaged F1 score.} \label{eval_res}
\centering
\renewcommand\cellalign{c}
\renewcommand{\arraystretch}{1.2}
\setlength{\tabcolsep}{0.6em}
\setcellgapes{3.3pt}\makegapedcells 

\begin{tabular}{l | c |  c | c | c || c |  c | c | c  }
\toprule
  &  \multicolumn{4}{c||}{\textbf{Column names}} &  \multicolumn{4}{c}{\textbf{Masked column names}} \\
 \cline{2-9}
 & $q=1$ & $q=3$ &  $q=5$ &  $q=7$  & $q=1$ & $q=3$ & $q=5$ & $q=7$  \\
\hline
tf-idf & 0.56 & 0.56 & 0.54 & 0.54 & 0.41 & 0.45 & 0.51 & 0.55\\
spacy & 0.64 & 0.69 & 0.74 & 0.73 & 0.48 & 0.58 & 0.61 & 0.63\\
word2vec & 0.69 & 0.76 & 0.76 &0.78 & 0.61 &\textbf{ 0.77} & 0.76 &\textbf{ 0.80} \\
bert & \textbf{0.76} & \textbf{0.78} & \textbf{0.79} &\textbf{0.80} & \textbf{0.63}& 0.75  & \textbf{0.78} & 0.78\\
tabert & 0.75 & 0.77 & 0.77 & 0.78 & 0.61 & 0.71 & 0.71 & 0.74\\
\toprule
\end{tabular}
\end{table}

\begin{figure}
  \begin{minipage}[b]{0.48\textwidth}
  \hspace*{-0.82cm}
    \includegraphics[width=\textwidth]{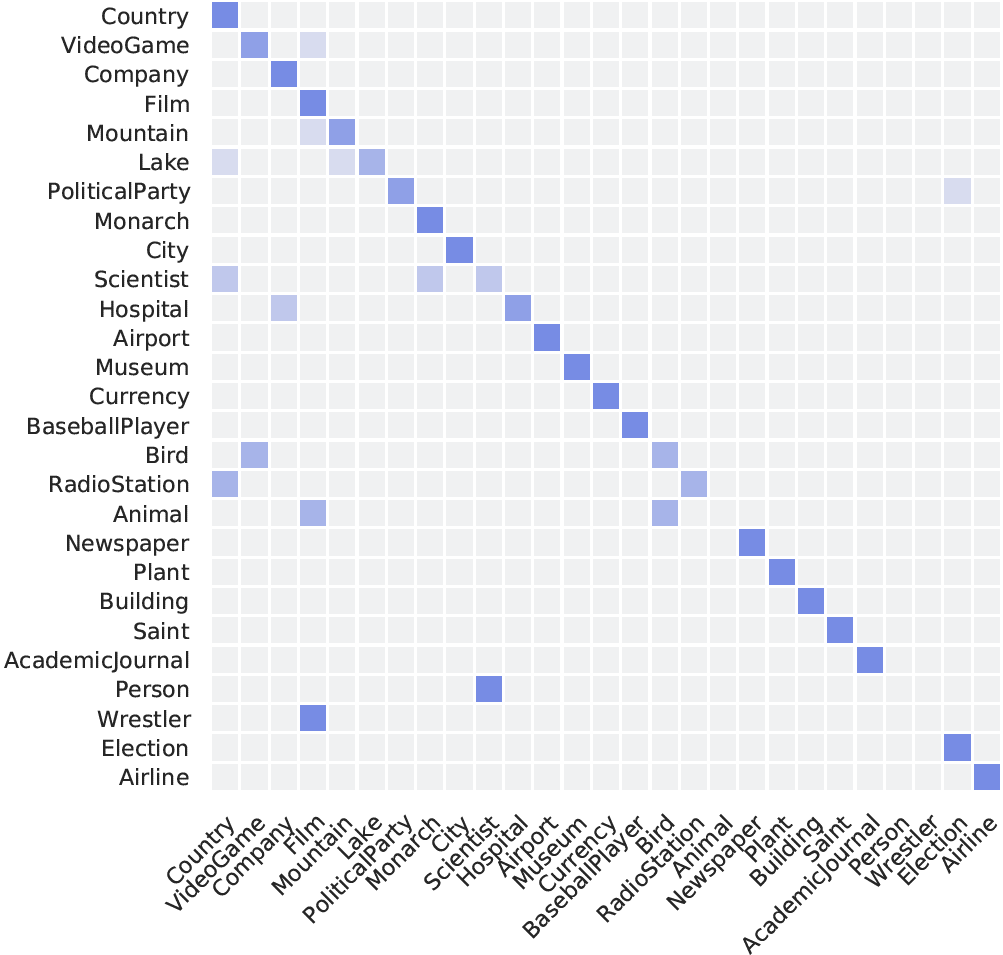}
  \end{minipage}
  \begin{minipage}[b]{0.48\textwidth}
  \raisebox{-0.5ex}
   {\hspace*{-0.9cm} \includegraphics[width=1.16\textwidth]{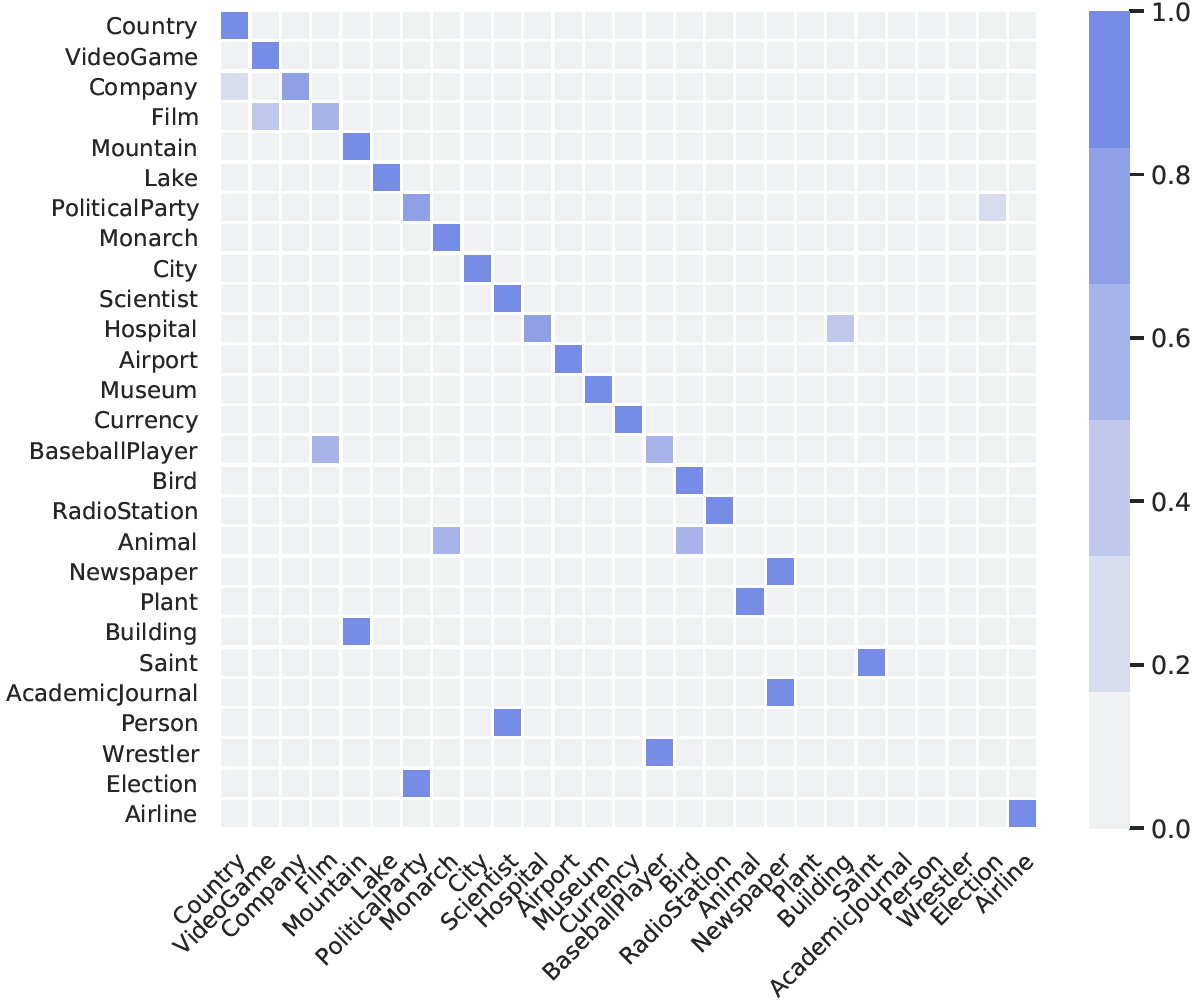} }
  \end{minipage}
  \caption{Classification confusion matrix for Word2Vec (left) and for TaBERT (right).} \label{conf_mtx}
 \end{figure}

Figure \ref{conf_mtx} shows the row-normalized confusion matrix for the table classification task for Word2Vec and TaBERT across the different classes. The horizontal axis shows the predicted labels and the vertical axis shows the true labels. 
We observe the performance of the two models under the same scenario: the input tables are with $q=7$ rows and the column names are masked. The classes are ordered by the number of instances assigned to them, \textit{Country} is the class with the most instances, $33$, while \textit{Airline} has only $2$ instances. From the confusion matrix for TaBERT (Figure \ref{conf_mtx} right) it can be observed that more miss-classifications are for the classes with a lower number of instances and they are not that unexpected. For instance, miss-classifying an instance of class \textit{Person} as an instance of class \textit{Scientist}, is an acceptable mistake. Similarly for the instances of classes \textit{Academic Journal} and \textit{Newspaper}, and \textit{Political Party} and \textit{Election}.
On the other hand, the miss-classifications by Word2Vec for class \textit{Wrestler} and class \textit{Animal} as instances of class \textit{Film} are much more unexpected and critical. 
Likewise, Word2Vec miss-classifies the tables of class \textit{Scientist} and of class \textit{Radio Station} as instances of the class \textit{Country} which indicates a weak semantic structure in the vector representations. These results suggest that although Word2Vec achieves higher F$1$ score, the TaBERT vector representations capture semantics with a smoother transitions between classes.

\begin{figure}[ht!]
    \centering
    \includegraphics[width=0.88\textwidth]{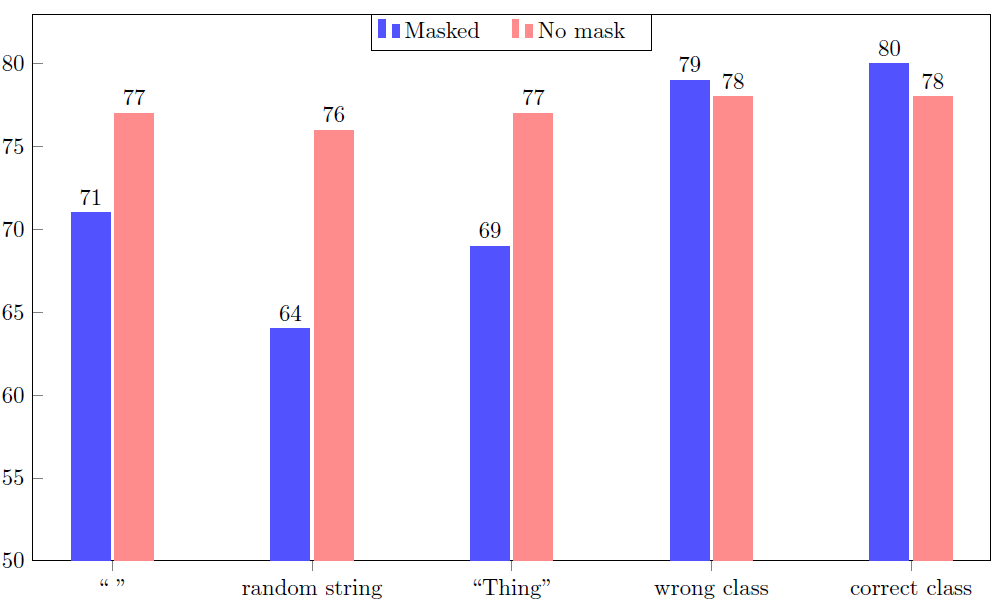}
    \caption{TaBERT performance with different utterances.}\label{tabert_context}
\end{figure}

\paragraph{\textbf{TaBERT Analysis}} To get a better understanding of the (under-) performance of TaBERT we analyse the influence of the \textit{utterance} and its interplay with column names. In addition to the empty string \textquote{ } used in previous experiments, we also used a randomly generated string with $10$ characters (unique per table), and one constant string, \textit{Thing}, for all tables. Moreover, we experimented with adding the correct class of the tables as utterance, as well as a wrong class (for instance, all the tables of class \textit{Country} are encoded with the class \textit{Plant} as utterance).
Figure \ref{tabert_context} shows the results of these experiments, where the input tables were with $q=3$ rows. The horizontal axis shows the different options that we passed as utterance to the model and the vertical axis shows the achieved F$1$ score. The masking of column names has significant influence on the generated table representation. The reason for this might be in the way how a row is transformed into a string, i.e., the value of each table entry is concatenated with the column name of the entry and its value. Observing the results with the different utterance, we see that the choice of utterance does not affect the performance of the model when the column names are not masked. Nevertheless, when the column names are masked, the influence of the utterance is more significant. In both cases when the utterance is the wrong class or the correct class, the achieved score is much higher, which might be attributed to a class-wide shift in the vector space because of the grouping that these utterances cause.
  
\vspace{-0.2cm}
\section{Conclusion and Future work}
In this paper we explored different types of table encoders for generating vector representations for tabular data. Specifically, we focused on evaluating different methods for table encoding on the sub-task for TI, table-to-class annotation. Despite the increasing interest in the problem of TI, so far, only one approach towards this specific sub-task has been proposed.
In this direction, we provided a formal definition for the table-to-class annotation task as a machine learning task.
We conduct an empirical study with five different methods for generating vector representation of a table and evaluate their performance on the table-to-class annotation task. 
The results from our experiments show that transfer learning methods with large vocabularies of pre-trained word embeddings perform on par with more complex and expensive modes such as LM pre-trained on tables. An interesting finding is that the inductive bias for tabular structure in TaBERT did not bring benefit to the performance of the BERT model. A possible explanation for this is the missing significant utterance that the TaBERT model expects as input. Nonetheless, the miss-classifications made by this model are reasonable, suggesting that the vector representations capture the semantics of the tables. 
Future work should target closing the gap between existing general-purpose models and model specific for encoding tabular data. 
To further our work we plan to explore other existing methods for table encoding for solving the table-to-class task, as well as for solving the entity-to-row and column-to-property tasks. 

\bibliography{sample-ceur}

\end{document}